\RequirePackage{luatex85}
\documentclass{article}

\usepackage{microtype}
\usepackage{graphicx}
\usepackage{subcaption}
\usepackage{booktabs} 

\usepackage[utf8]{inputenc}
\usepackage[T1]{fontenc} 
\usepackage{url} 
\usepackage{amsfonts}
\usepackage{nicefrac}       
\usepackage{microtype}      
\usepackage{xcolor}         
\usepackage{enumitem}
\usepackage{array}
\usepackage{tablefootnote}
\usepackage{subcaption}
\usepackage{ragged2e}
\usepackage{pifont}
\usepackage{tabularx}
\usepackage{multirow}
\usepackage{multicol}
\usepackage{csquotes}

\usepackage{arydshln}

\newcolumntype{P}[1]{>{\centering\arraybackslash}p{#1}}

\newcolumntype{P}[1]{>{\RaggedRight\arraybackslash}p{#1}}

\usepackage{hyperref}

\usepackage[preprint]{icml2026}

\usepackage{amsmath}
\usepackage{amssymb}
\usepackage{mathtools}
\usepackage{amsthm}

\usepackage{makecell}

\usepackage[capitalize,noabbrev]{cleveref}

\theoremstyle{plain}

\theoremstyle{definition}

\theoremstyle{remark}

\usepackage[textsize=tiny]{todonotes}

\usepackage{amsmath}
\usepackage{subcaption}
\usepackage{booktabs} 
\usepackage{tabularx} 
\usepackage{arydshln}
\usepackage{ragged2e}
\usepackage{multirow}
\usepackage{multicol}

\usepackage[most]{tcolorbox}
\usepackage{booktabs}
\usepackage{amsmath}

\icmltitlerunning{On the Diversity of Analogy Making in Large Language Models}

\begin{document}

\twocolumn[
  \icmltitle{On the Diversity of Analogy Making in Large Language Models}



  \icmlsetsymbol{equal}{*}

  \begin{icmlauthorlist}
    \icmlauthor{Yuanhao Shen}{queen}
    \icmlauthor{Daniel Xavier de Sousa}{brazil}
    \icmlauthor{Caio César Sifuentes Barcelos}{brazil}\\
    \icmlauthor{Hongyu Guo}{nrc}
    \icmlauthor{Xiaodan Zhu}{queen}
  \end{icmlauthorlist}

  \icmlaffiliation{queen}{Department of Electrical and Computer Engineering \& Ingenuity Labs Research Institute, Queen's University, Canada}
  \icmlaffiliation{nrc}{National Research Council Canada}
  \icmlaffiliation{brazil}{Instituto Federal de Goiás, Anápolis, Brazil}

  \icmlcorrespondingauthor{Yuanhao Shen}{23rq31@queensu.ca}

  \icmlkeywords{Machine Learning, ICML}

  \vskip 0.3in
]






\printAffiliationsAndNotice{}  

\begin{abstract}



Large Language Models (LLMs) have demonstrated remarkable potential for analogy making, a core cognitive capability that drives novelty and creativity. 
While prior research has extensively investigated the applications and underlying mechanisms of LLM-based analogy making, its output diversity remains largely unexplored, despite being essential for broadening cross-domain connections and fostering scientific innovation. 
In this work, we present a comprehensive evaluation of analogy diversity across ten state-of-the-art open- and closed-source LLMs. 
Our findings highlight a  concerning issue of domain homogeneity, a prevalent tendency for LLMs to generate analogies from a narrow set of target domains, limiting both inter-query and intra-model diversity. 
Furthermore, our analysis reveals a fundamental trade-off in existing LLM diversity-enhancement methods: increasing output diversity often comes at the expense of output quality. Finally, our causal analysis of LLM internals uncovers substantial differences in the model-sensitive regions governing analogy diversity across LLMs, suggesting a potential mechanistic basis for the observed diversity–quality trade-off. To our knowledge, this is among the first studies to systematically investigate output diversity in LLM-based analogy making. Our findings provide empirical insights and practical guidance for developing more diverse LLMs for analogy generation. Our code are available at \texttt{[url\_placeholder]}.

\end{abstract}

\vspace*{-10mm}
\section{Introduction}
\label{sec:intro}





\noindent 




Recent advances in Large Language Models (LLMs) have demonstrated remarkable potential to accelerate scientific discovery, ushering in a new era of AI-for-Science (AI4S). 
Beyond their broad range of applications in domains such as protein design \cite{Fry2026_protein, Liu2025_protein}, sustainable energy \cite{Hong2026_energy, zhang2026machine_energy}, materials science \cite{park2026guiding}, and interdisciplinary knowledge discovery \cite{IDRBench26}, recent research has begun to explore LLMs' emerging capacity to generate diverse and creative analogies across scientific domains, a cognitive capability central to human intelligence that sparks scientific innovation~\cite{shen2026unlocking}.

\begin{figure}
    \centering
    \includegraphics[width=0.85\linewidth]{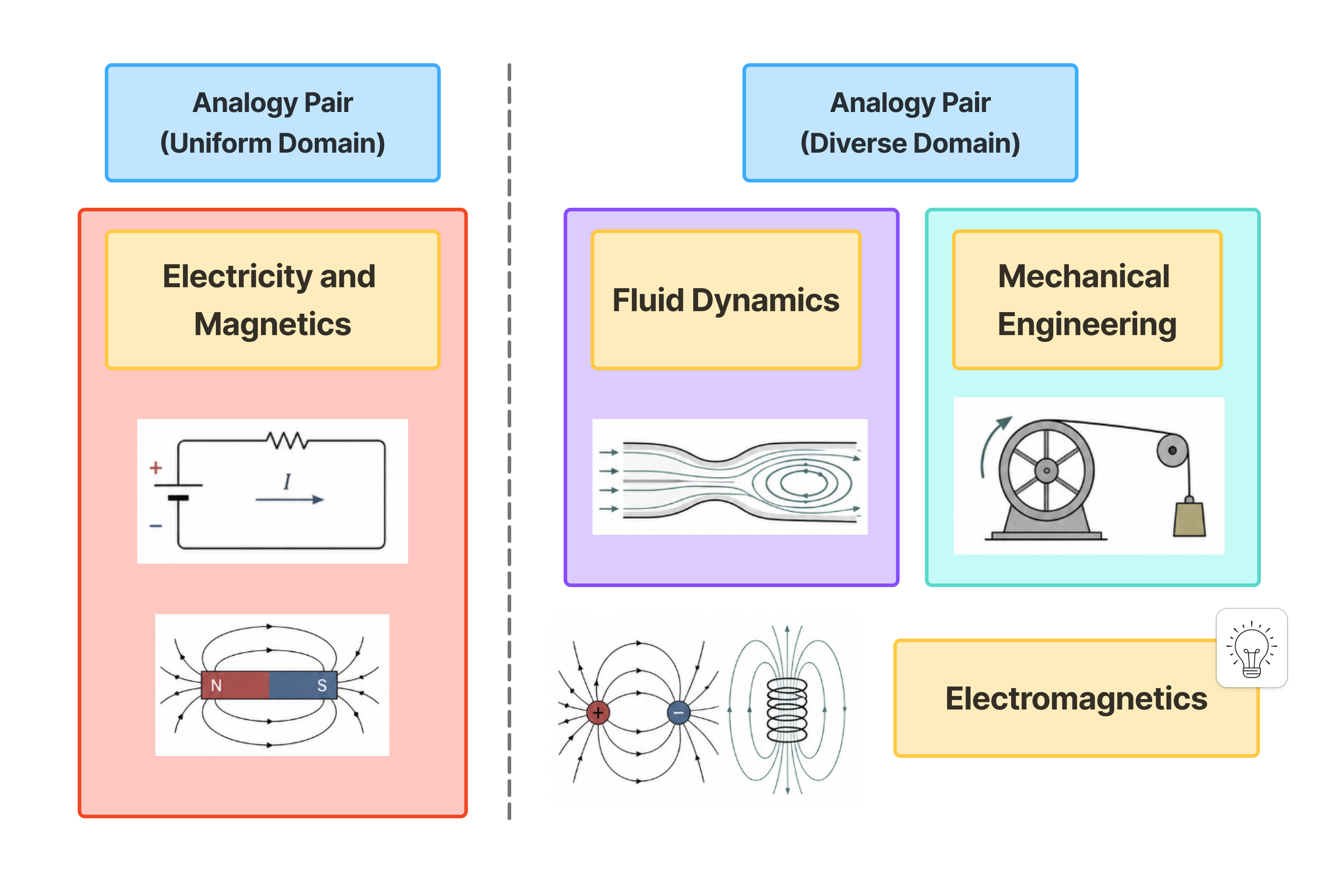}
    \caption{
    Maxwell's development of the theory of electromagnetism through analogical reasoning across diverse domains. The left subfigure illustrates analogies drawn from closely related domains, while the right subfigure shows more diverse cross-domain analogies. This example demonstrates how greater analogy diversity enables broader knowledge integration and can foster scientific discovery, motivating our study of analogy diversity in large language models.}
    \label{fig:motivation_example}
    \vspace{-3mm}
\end{figure}

In AI4S, where scientific innovation is driven by interdisciplinary knowledge integration, analogy diversity strengthens LLMs' ability to uncover both explicit and implicit connections across domains and disciplines~\cite{IDRBench26}. 
A classic historical example is James Clerk Maxwell's development of electromagnetic theory~\cite{clerk1864faraday}. Maxwell employed fluid and mechanical systems as the \textbf{source domain} to reason about electromagnetic phenomena, the \textbf{target domain}. Although these domains are conceptually distant, their structural correspondence enabled a breakthrough in scientific understanding, highlighting the importance of diverse analogies in driving innovation, as illustrated in Figure~\ref{fig:motivation_example}.  

Motivated by this observation, a fundamental question is whether current LLMs can generate the diverse analogies needed to foster cross-domain knowledge transfer and scientific innovation. Although LLM-based analogy making has received increasing attention, with existing research primarily focusing on analogy quality, applications, and underlying mechanisms \cite{shen2026unlocking}, the diversity of generated analogies remains largely unexplored. Different from conventional generation tasks, analogy-making explicitly requires a departure from the source domain and a bold selection of the target domain, where both source and target domains share similar functional abstractions \cite{Analogy83, domain_function}. The overlook of diversity in analogy making further hinders a comprehensive understanding in LLM analogy-making mechanisms, leaving a notable gap to this central research question: \textbf{Are LLMs capable of \textit{generating} diverse and informative analogies?}


\begin{figure*}[t]
    \centering
    \includegraphics[width=0.86\linewidth]{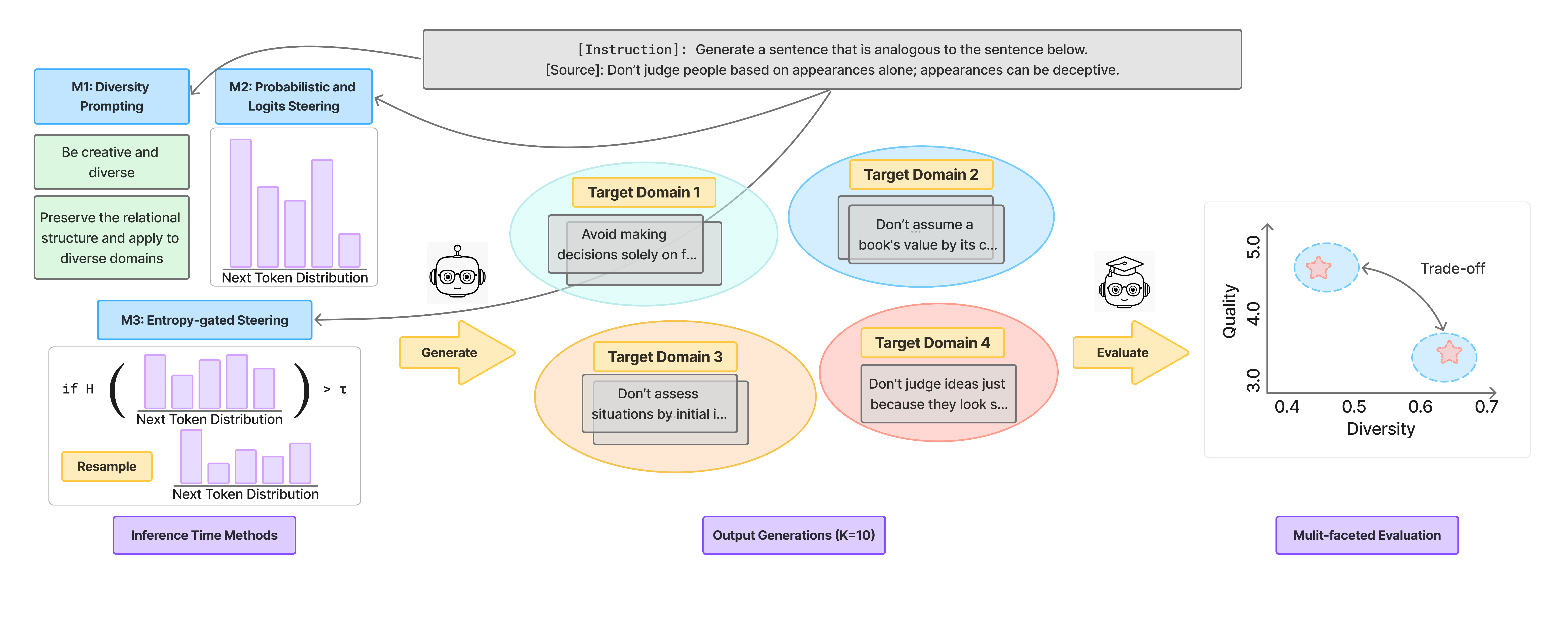}
    \caption{Overview of experiments and evaluation methods on LLM's analogy making diversity. 
    The model is given the input of the format \texttt{[Instruction, source]} along with different inference time methods in the left column. We then cluster the output generations into different target domains and conduct multi-faceted evaluations on those generations.}
    
    \label{fig:overview}
\end{figure*}

In this work, we provide a thorough evaluation of analogy-making diversity across multiple LLMs from two complementary perspectives. First, we evaluate how well mainstream LLMs are able to generate analogical sentences by introducing novel and multi-faceted metrics that focus on domain selection diversity. Our experimental results show that, LLMs tend to generate analogous sentences closely aligned with similar target domains, exhibiting a homogeneity trend across different model families. The excessive homogeneity in analogy generation poses a fundamental limitation when these models are applied to scientific discovery, a process that inherently relies on diversity and creativity.

Second, to better understand the effect of existing diversity-enhancing methods under an analogy making setting, we evaluate approaches that span three categories of inference-time perturbation, including (i) diversity prompting based methods, (ii) probability and logits steering, and (iii) entropy-gated steering. Our experimental results show that existing methods provide only limited improvements in analogy diversity and are often accompanied by the sacrifice of analogy quality. The trade-off between analogy diversity and output quality suggests that techniques originally designed for general-purpose generation may not readily address domain homogeneity in analogy-making tasks. Additional analysis on LLMs' internal layers further suggests that such inefficacy might be caused by a lack of a consistent region of analogy making across different models.


To the best of our knowledge, our work is among the first to systematically study output diversity under the context of LLM analogy generation. We summarize our  main contributions as follows:
\begin{itemize}
    \item \textbf{Comprehensive evaluation of LLM analogy making diversity.} Our work provides a thorough evaluation of diversity in LLMs' analogy making, using a novel and multi-faceted metrics that provide detailed insights into the output of LLM generations.
    
     \item \textbf{Domain homogeneity in LLM analogy making.} Our results reveal domain homogeneity in analogy making of these models based on both intra-query and intra-model perspective, highlighting an alarming limitation in the promotion of creativity and diversity when applying LLMs in scientific innovation.
     \item \textbf{Diversity-quality tradeoff in LLM analogy making.} Further analysis on the efficacy of applying diversity enhancing frameworks in LLM analogy making shows a trade-off between output diversity and quality. We also identify notable differences in perturbation sensitivity in different models, showing a divergence in models' internal behaviors in analogy making.
\end{itemize}



\section{Related Works}



\paragraph{Diversity in Generation.} \hspace{-2mm} Generation diversity in Large Language Models (LLMs) has been well-studied, especially in the realm of applications in creative writing \cite{chung2025diversityincreativewriting}. Prior work has demonstrated the importance of diversity in applications such as short story completion~\cite{tian-etal-2024-large-language} and creative writing~\cite{minkSampling26}, where creative variation and exploration of alternative narratives are fundamental to high-quality generation. To improve diversity, early works aim to solve the mode collapse phenomenon in LLMs \cite{jiang2025modecollapse}, where models often generate repetitive, bland, or generic outputs. Moreover, frameworks such as beam search \cite{wu2016beamsearch}, Top-$k$ and Nucleus (Top-$p$) sampling \cite{fan2018topk, holtzman2020topp} are proposed to mitigate the issue. Beyond inference-time strategies to promote diversity, metrics such as \textit{Distinct-$n$} \cite{li2016distinctn} and \textit{Self-BLEU} \cite{zhu2018selfbleu} have become standard for measuring distributional breadth, while recent efforts have begun to emphasize "effective semantic diversity" to ensure that variety does not come at the cost of task-specific quality \cite{wiher2022decoding}. However, these works often address surface-level or stylistic diversity in creative writing, leaving a sparse space for the exploration of outcome diversity in cognitive tasks such as analogical reasoning.


\paragraph{Analogical Reasoning in LLMs.} \hspace{-2mm}
Analogical reasoning is deemed a central faculty of human intelligence \cite{minegishi2026arintransformers, hofstadter2013surfaces}. Modern methods that study human analogy often refer to the process of abstraction and structure mapping that make hops across different domains \cite{getner2010bootstrapping}. The recent burgeoning of mechanistic interpretability methods \cite{huben2024sparse, Belrose2023ElicitingLP, ghandeharioun2024patchscopes} has fueled studies on LLM analogical reasoning abilities from an anatomical perspective, offering a shift towards a more comprehensive understanding within LLMs than of their human counterparts. In light of pioneering works that treat analogy-making as a dual interplay between domain and function \cite{Analogy83}, recent works make remarkable progress ranging from model behavioral studies \cite{webb2023emergent} to efforts that explore underlying mechanisms, with proposals of certain hypotheses such as structure mapping \cite{Analogy83} and graph functor \cite{minegishi2026arintransformers}. However, to our knowledge, all of these explorations are conducted under a classification setup, which often involves comparisons between analogous pairs, thus causing these efforts either to remain relatively superficial or to bear the risk of obliterating the expansive nature of the solution space in analogy-making, as argued in the book \citet{hofstadter2013surfaces}. To bridge this gap, our work moves beyond comparing binary labels to the open-ended generation analysis in LLM analogical reasoning.





\section{Experiment Setup}
\label{sec:exp_setup}

\paragraph{Datasets.} \hspace{-2mm} To provide a comprehensive evaluation of diversity in LLM-based analogy, we conduct experiments on three recently released analogy generation datasets. AnaloBench (AB) \cite{analobench} is a benchmark dedicated to story-level analogy evaluation that contains 340 pairs of high-quality analogous stories from human annotators. Metaphoric Analogies (MA) \cite{boisson-etal-2025-automatic} focuses on extracting structured analogical mappings from literary metaphors with 203 samples that build source and target concept pairs in a four-term analogy structure. Metaphor Understanding Challenge (MUNCH) \cite{tong-etal-2024-metaphor} evaluates whether LLMs can understand metaphors as cross-domain mappings through paraphrasing, providing over 10K paraphrases for sentences containing metaphor use. We use the full evaluation dataset for both AB and MA; for MUNCH, we extract 1K samples randomly. 

\paragraph{Models and Devices.} \hspace{-2mm} We include five closed-source and five open-source models in our evaluation. We use \texttt{GPT-5.2}, \texttt{Grok-4.5}, \texttt{gemini-2.5-pro}, \texttt{gemini-3.1-pro}, and \texttt{claude-4.6-sonnet} for generation diversity evaluation. We further test five open source models, including \texttt{llama-3.1-8B-instruct}, \texttt{gemma2-9B-it}, \texttt{qwen-3-8B}, \texttt{mistral-7B-Instruct}, and \texttt{phi-4-mini-instruct}. The maximum generation length is set to 100 output tokens for all open source models. We run our experiments using single \texttt{l40s} and \texttt{h100} GPUs. 

\paragraph{Generation Perturbation Methods.} \hspace{-2mm} We evaluate a spectrum of inference-time perturbation methods following the categories defined in \citet{ostermann-etal-2026-weights} to provide a comprehensive understanding of their efficacy in LLM analogy making.


\begin{itemize}
    \item \textbf{Diversity Prompting.} In addition to base prompting, we run diversity prompting strategies, including Logic prompts and Diversity prompts to explicitly encourage analogies from creative and diverse domains.
    
    \item \textbf{Probabilistic and Logits Steering.} We run probabilistic methods including Top-$k$ ($k=20$), Top-$p$ ($p=0.9$), Top-$\eta\sigma$ ($\eta\sigma=1.0$), and Min-$p$ ($p=0.1$) in our experiments that resample the output logits to improve output diversity. We also include G2 \cite{G2_25} in our experiments, where we set the number of iteration rounds to be 3.
    \item \textbf{Entropy-gated Steering.} We also include an adaptation from \citet{li-etal-2026-entropy} that uses entropy as an indicator to control the output diversity. We set the resampling threshold $\tau$ to be larger than the 90th percentile of the whole generated token sequence. 
    
\end{itemize}


\paragraph{Output Evaluations.} \hspace{-2mm} We run our evaluation from $K = 10$ generations for each sample. Following previous studies that evaluate generation diversity by mapping natural language into a semantic space \cite{reimers2019sentencebert}, we adopt a similar paradigm. Specifically, we employ the \texttt{all-MiniLM-L6-v2} sentence transformer as an encoder to obtain semantic embeddings, calculating their mean pairwise cosine distance. Furthermore, we incorporate the MAUVE score \cite{pillutla2022mauve} to better capture distributional differences, providing a more comprehensive analysis of generation over quality and diversity. To evaluate the coherence and quality of the generated analogies, we use \texttt{GPT-4o-mini} under an existing LLM-as-a-judge setup \cite{zheng2023llmasajudge}. 



\begin{figure*}[t]
    \centering
    \begin{minipage}[b]{0.66\textwidth}
        \centering
        \includegraphics[width=\linewidth]{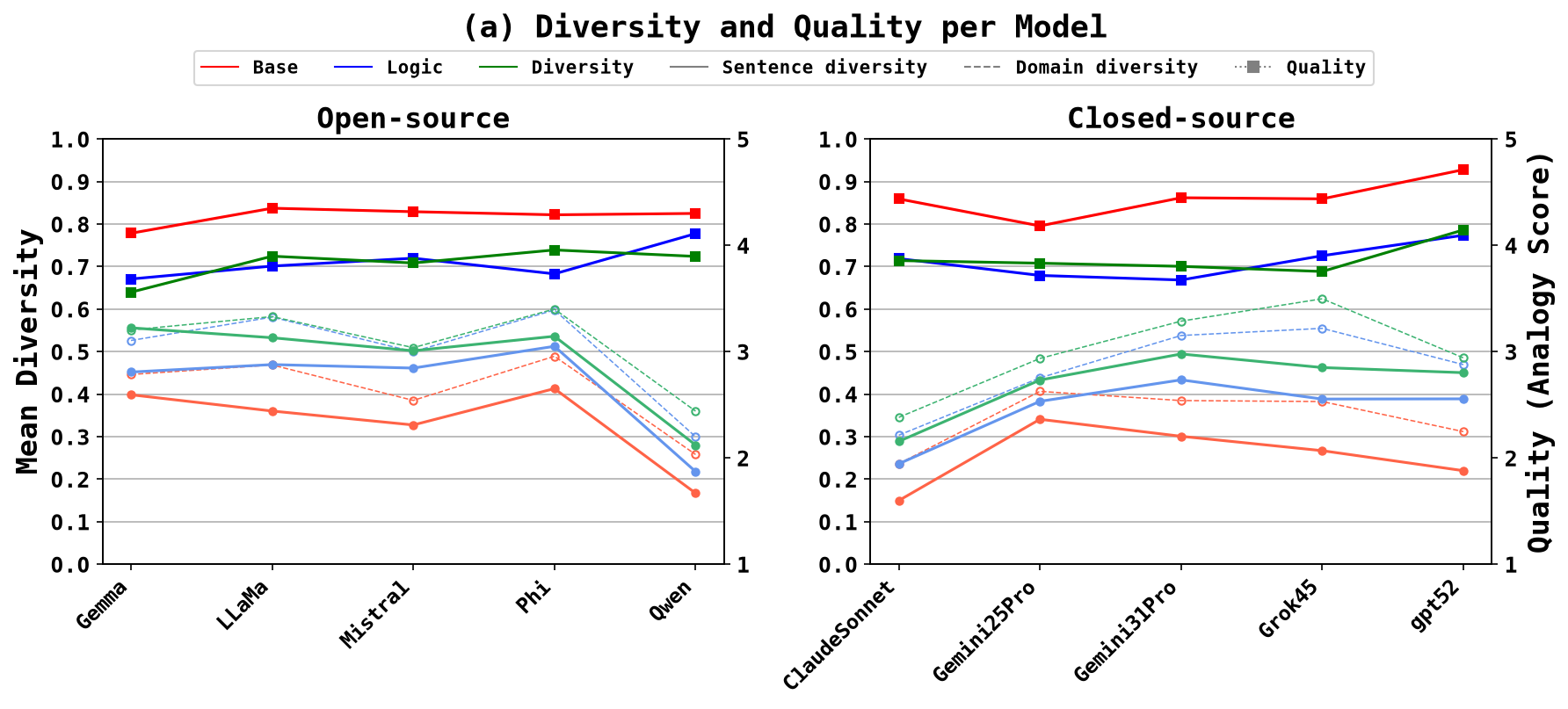}
    \end{minipage}
    \hfill
    \begin{minipage}[b]{0.33\textwidth}
        \centering
        \includegraphics[width=\linewidth]{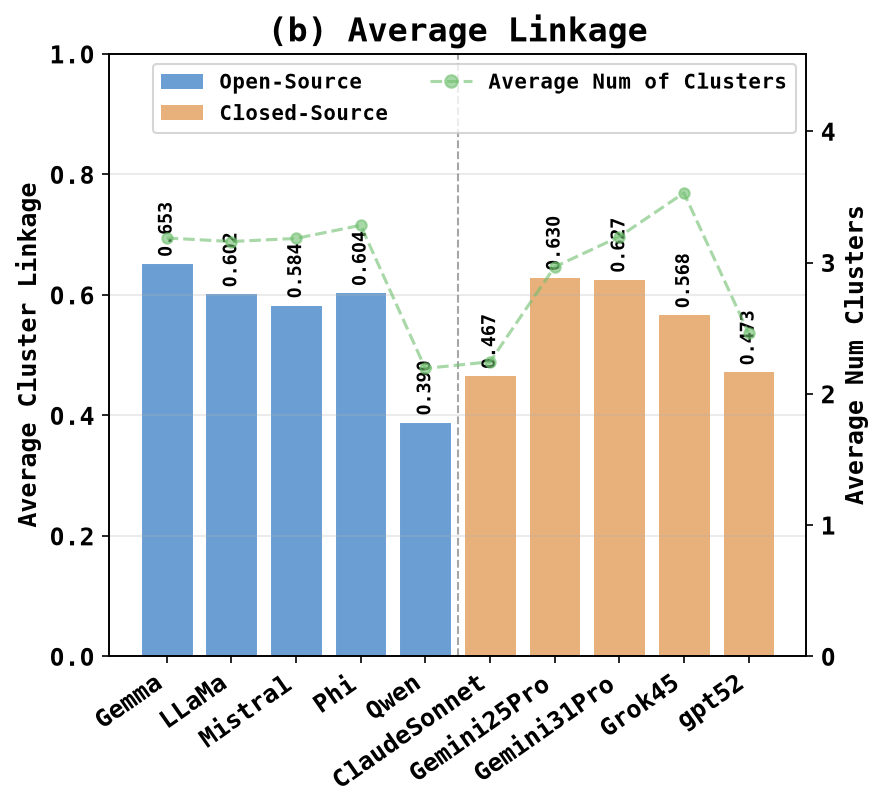}
    \end{minipage}
    \vspace*{-5mm}
    \caption{(a) Comparison of sentence diversity, domain diversity, and output quality across prompting strategies. Solid lines denote sentence diversity, dashed lines denote domain diversity, and square markers indicate quality scores for open-source and closed-source models. (b) Inter-domain diversity using average linkage. Bars denote average linkage; dashed line indicates mean clusters per sample.}
    \label{fig:div_analysis}
\end{figure*}

\section{Domain Homogeneity in Analogy Generation}
\label{sec:motivation}


We first conduct a study in diversity of models' analogy-generation capabilities, focusing on whether the generated analogous sentences fall into diverse target domains. We assess this from two perspectives: i) Intra-output analogy diversity, which measures the pairwise and clustering distances among generations within a single model, and ii) Inter-model analogy diversity measures, which assess whether analogy generation converges to the same target domain or commonalities in sentence structure across different models. 


\begin{figure}[ht]
    \begin{center}
    \hspace*{-8mm}\includegraphics[width=0.75\linewidth]{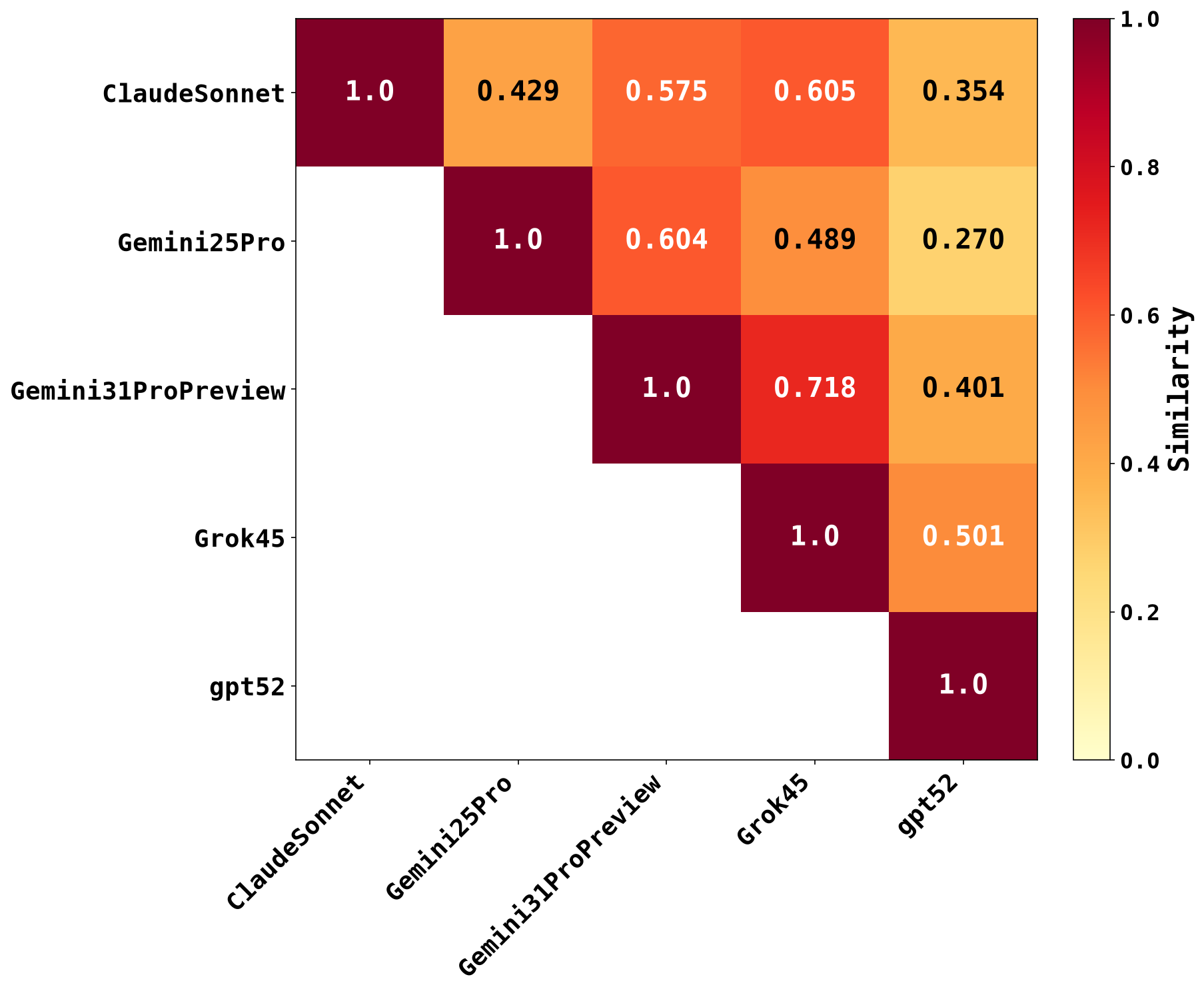}
    \end{center}
    \vspace*{-5mm}
    \caption{Heatmap for Inter-Model MAUVE scores. MAUVE is originally designed to measure how closely machine-generated text matches human-written text. We adapt MAUVE to compare how closely related the generated texts are across different close source models.}
    \label{fig:mauve_comp}
    \vspace*{-5mm}
\end{figure}

\subsection{Intra-Output Diversity}
\label{subsec:intra_output_diversity}
To obtain intra-output diversity for each specific model, following similar evaluation for general diversity~\cite{G2_25, minkSampling26}, we first measure the pairwise distances among the distinct generations to show how closely related each generation is to the others. Specifically, we assemble prompts in the form of a binary tuple \texttt{[Instruction, Source]}, where the \texttt{Instruction} field asks the model to generate analogical sentences under \textit{base}, \textit{diverse}, or \textit{logical} prompting strategies, and the \texttt{Source} field contains a sentence where the model should identify its source domain and generate analogies thereafter. Under the \textit{base} setting, the model is simply instructed to generate an analogy. In the \textit{diverse} setting, the model is additionally encouraged to produce creative and diverse outputs. Finally, in the \textit{logical} setting, the model is instructed to preserve the same relational structure while generating analogies in different target domains, without explicitly encouraging diversity. 

Figure \ref{fig:div_analysis} (a) presents the pairwise cosine distances between generation embeddings from two perspectives: (i) solid lines (less saturated colors), where embeddings are computed from the complete generated analogies, and (ii) dashed lines, where embeddings are computed only from the extracted words describing the target domains. For the latter, target domains are extracted using \texttt{GPT-4o-mini}. To evaluate the quality of the generated analogies, following \citet{zheng2023llmasajudge}, we employ an LLM-as-a-judge approach, with the results also reported in the figure with solid lines and saturated colors.

Our results show intuitive behavior of the tested models: when explicitly instructed to be diverse, the models generate more creative outputs, while prompts emphasizing logical behavior also improve diversity compared to naive prompting strategies (less saturated colors in Figure \ref{fig:div_analysis} (a). However, the quality tends to be better for base prompts (red saturated color). Comparing closed-source and open-source models, there is no clear evidence of superiority between them, showing that more complex post-training may not necessarily improve diversity. Nevertheless, even among the mainstream closed-source models evaluated, the best-performing model for the diversity prompt (less saturated solid line in the figure), \texttt{Gemini3.1-Pro}, achieves only an average intra-output diversity of 0.49, indicating that overall diversity remains limited. Considering only the extracted target domains, where irrelevant lexical content is removed and the evaluation focuses solely on the target domains, all models exhibit improved diversity, showing a similar standard among the prompts. Nevertheless, the overall diversity remains limited, with the mean pairwise diversity not exceeding 0.63.

Except for the target domain extraction step, the evaluation shown in Figure \ref{fig:div_analysis} (a) follows a standard approach described in several works for assessing diversity in text generation. However, a key limitation of this evaluation is that it remains difficult to interpret whether an average similarity score of 0.6 truly reflects high or low diversity. To provide a complementary perspective, we employ \texttt{GPT-4o-mini} to identify the semantic target domains of the generated analogies and group the outputs into distinct clusters. These clusters enable us to assess diversity at the semantic level by considering both the number of discovered clusters and the hierarchical Average Linkage, i.e., the average similarity over all pairs of sentences belonging to different clusters \footnote{We also evaluated Single Linkage and Complete Linkage and obtained similar results.}, where a greater number of clusters and higher Average Linkage values indicate greater diversity. Similar clustering-based strategies have been adopted in other contexts to evaluate output diversity \cite{pillutla2022mauve}.

Figure \ref{fig:div_analysis} (b) presents the clustering results obtained using the diverse prompt across all evaluated models. Despite explicitly instructing the models to generate 10 distinct and diverse analogies, we observe an average of only two semantic clusters per model. Among the evaluated models, \texttt{Grok-4.5} produces the highest diversity, with an average of 3.5 clusters, whereas \texttt{Qwen3-8B} generates fewer than two clusters on average. Regarding the linkage analysis, both closed-source and open-source models exhibit comparable Average Linkage values (0.39--0.65), indicating similar levels of semantic separation between target domains. With an average of three clusters, \texttt{Gemma-2-9B} achieves the highest Average Linkage ($\overline{\text{avg}} = 0.653$), followed by the Gemini family among the closed-source models ($\overline{\text{avg}} = 0.628$). In contrast, \texttt{Qwen3-8B} obtains the lowest score ($\overline{\text{avg}} = 0.390$). To assess the reliability of the results in Figure \ref{fig:div_analysis}, we conducted a human evaluation of the generated clusters. Based on a randomly sampled subset comprising 10\% of the data, the human annotations agreed with the automatic clustering in 95\% of the cases. 

These findings reveal a notable tendency toward target-domain homogeneity in analogy generation, raising concerns about the capability of current LLMs to consistently produce diverse analogical mappings. For instance, when the model is provided with a \texttt{context} sentence starting with "Don't judge a person...", the model has strong preferences for repeating similar target domains such as "book" or "quiet river", with the variety not following a uniform distribution for $K$ analogy generations.

\subsection{Inter-Model Diversity}
In addition to domain homogeneity within a single model, we further explore the diversity across different models. Using the diverse prompt settings of the previous section, Figure \ref{fig:mauve_comp} showcases the inter-model diversity comparison heatmap reported using adapted MAUVE scores \cite{pillutla2022mauve}, where a higher score means more similar and less diverse. As described in Figure \ref{fig:mauve_comp}, although some model pairs exhibit different domain selection preferences (\texttt{gemini-2.5-pro} v.s. both \texttt{gpt-5.2} and \texttt{claude-4.6-sonnet}), the inter-model MAUVE scores remain high (less diverse) for the remaining models in the comparison. Notably, when comparing the generation distribution between \texttt{grok-4.5} and \texttt{gemini-3.1-pro-preview}, we obtain a MAUVE score of 0.72, indicating an overwhelming overlap in the generation patterns between these models. 

\subsection{Data Pollution}
\begin{figure}[ht]
    \begin{center}
    \hspace*{-8mm}\includegraphics[width=0.75\linewidth]{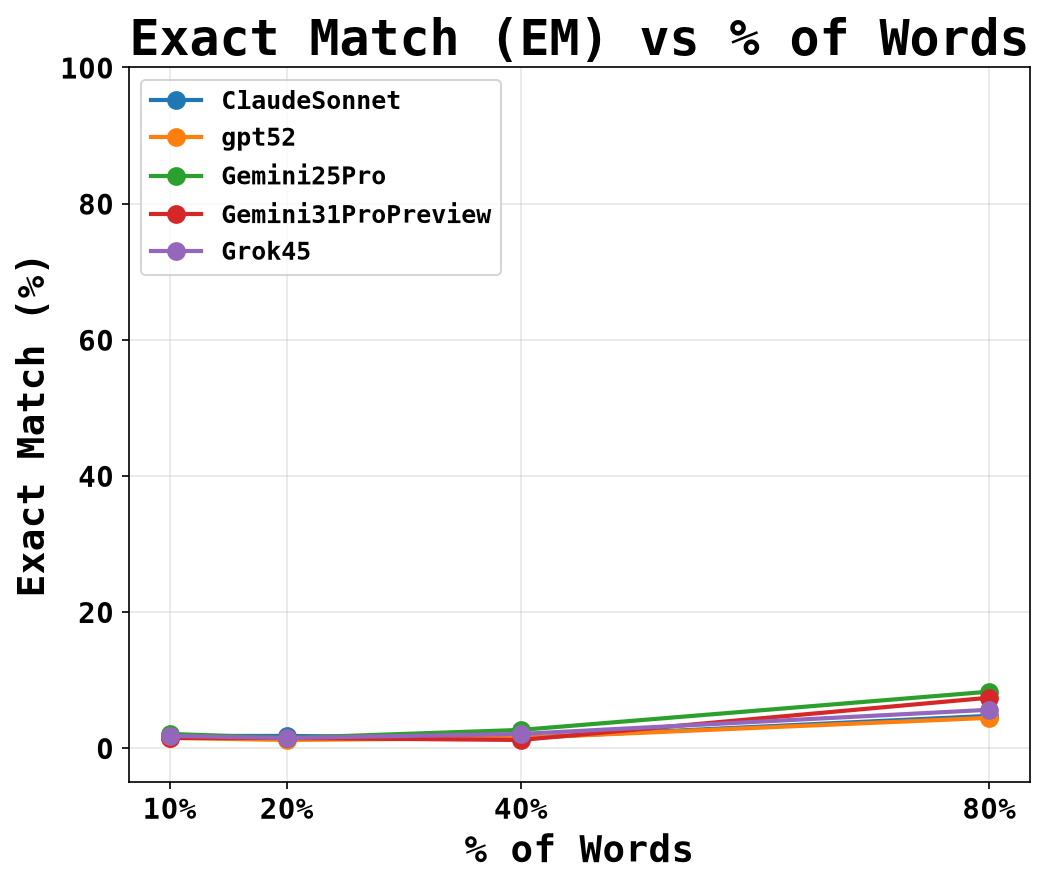}
    \end{center}
    \vspace*{-5mm}
    \caption{Analysis on data pollution. We use the method from \citet{golchin2024time} that makes varied percentage of context and compute the recovery accuracy with the original sentence.}
    \label{fig:data_pollution}
    \vspace*{-5mm}
\end{figure}

One potential issue that poses a threat to the accuracy of our comparison is data pollution, i.e., the model has already memorized the corresponding analogous pairs given the \texttt{source}. To tackle this challenge, we apply the method introduced by \citet{golchin2024time} to ensure there is no data pollution issue. Specifically, for each sample in our dataset, we mask out words at different percentage levels and check whether the model is able to recover the complete sentence measured by token exact match. Figure \ref{fig:data_pollution} reports the results in our analysis. It can be seen that all the models in our experiment are \textbf{not} suffering from a data pollution issue, as the highest exact match score of 14\% is achieved by \texttt{gemini-3.1-pro-preview} at the 80\% context level. This means that even the best-performing model only recovers a small portion of the original sentence when 80\% of the context is provided. The analysis of data pollution further enhances the validity of our observation regarding domain homogeneity. 

\section{Evaluation of Diversity Enhancing Methods in Analogy Making}
\label{sec:results}

Given the domain homogeneity in LLM analogy making observed above, we further explore the efficacy of existing frameworks that aim to enhance the output diversity of LLMs. As is mentioned in \citet{ostermann-etal-2026-weights}, existing methods that improve output diversity can be divided into three categories, including fine-tuning, prompting, and steering. Since the scope of our work lies at inference time, we mainly investigate the latter two methods of prompting and steering. 

For prompting, we are using the best results from our previous section, \textit{basic} and \textit{diverse} settings, which are optimal in quality and diversity respectively. 
We also include approaches that fall into the steering category and tend to perturb the model's internal hidden states in order to modify and encourage output diversity. We include approaches that span two mainstream subcategories from this perspective in our evaluation, including probabilistic / logistic resampling and entropy-gated resampling. 

\paragraph{Probabilistic and Logistic steering.} \hspace{-2mm} We employ \textbf{Top-$k$} that restricts sampling to the $k$ most probable tokens \cite{fan2018topk}; \textbf{Top-$p$} that samples from the smallest set of tokens whose cumulative probability reaches a threshold \cite{holtzman2020topp}; \textbf{Top-$\eta\sigma$} that retains tokens whose logits are within $\eta$ standard deviations of the highest logit \cite{tang2024topes}, and \textbf{Min-$p$} that discards tokens whose probability falls below a minimum threshold relative to the most probable token \cite{minh2025minp}. We also include G2 \cite{G2_25} in our evaluation \footnote{We include a reproduction of G2 following the algorithm in the github repository.}. G2 modifies the output logits based on previous outputs from guide models to improve diversity.

\begin{figure*}[t]
    \centering
    \includegraphics[
        width=0.9\textwidth,
        height=0.3\textheight,
        keepaspectratio
    ]{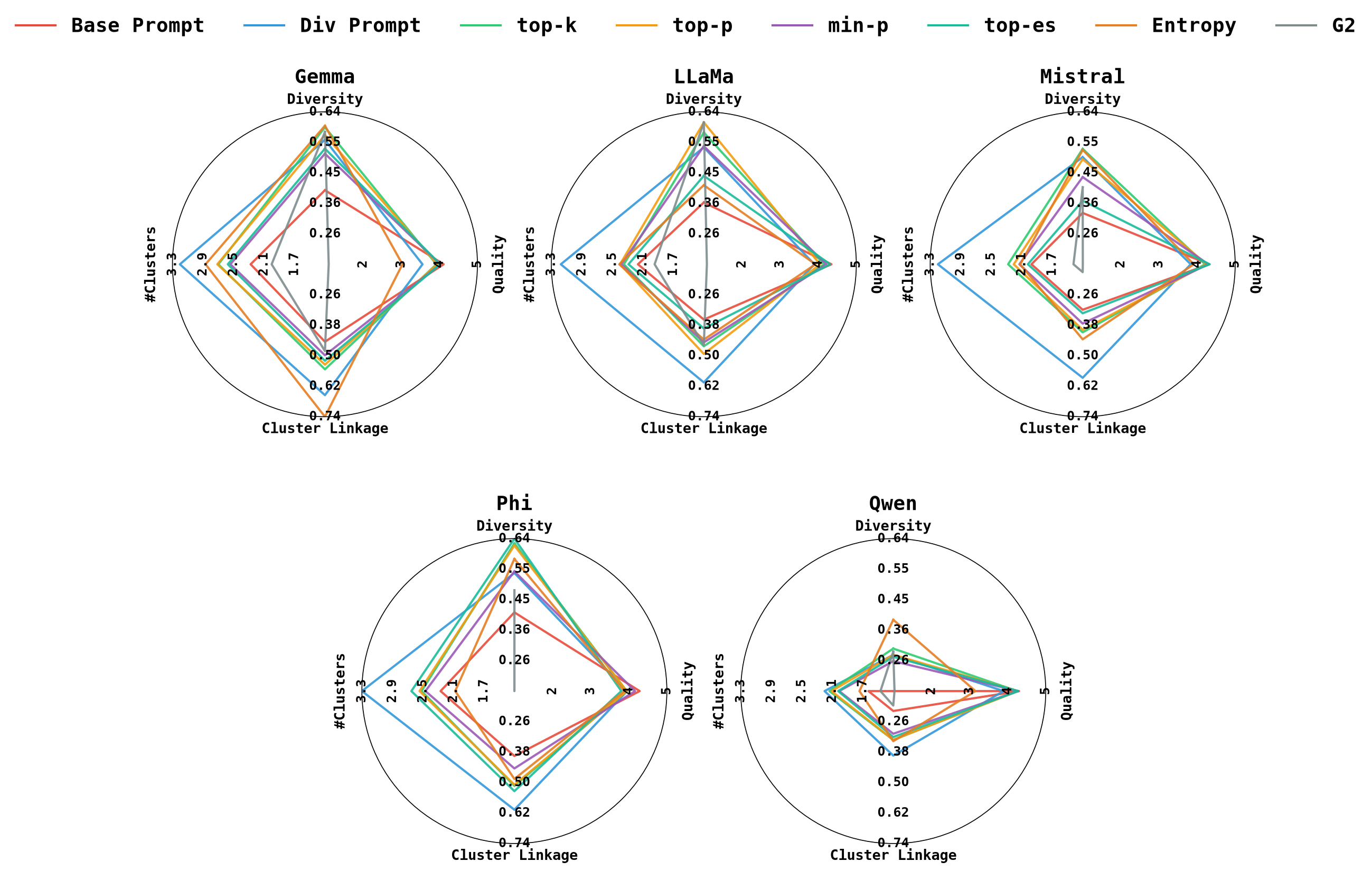}
    \caption{Radar chart comparing 8 decoding methods across 5 LLMs. Each model is shown in its own polar plot with 4 metrics (Diversity, Cluster Linkage, Quality, \#Clusters) normalized to 0,1 via min-max scaling (quality capped at 1,5). Tick labels on each spoke indicate the original-scale values.}
    \label{fig:radar_chart}
\end{figure*}

\paragraph{Entropy-gated steering.} \hspace{-2mm} We take the motivation and insights from \cite{li-etal-2026-entropy} and implement an adapted version to accommodate our task of analogy making, where we set a threshold $\tau$ with respect to the entropy of the next token to be generated. The resampling operation is only triggered when the entropy is higher than $\tau$ and the next token to be generated is identified as the domain token. Following observations in \cite{fan2026bridginggaplatentexplicit}, we run an additional forward pass in the same model in order to modify the output token distribution for resampling. The left column in Figure \ref{fig:overview} summarizes these methods in our evaluation. 

We showcase the performance comparison averaged across three datasets in a radar chart in Figure \ref{fig:radar_chart} \footnote{Complete results are also provided in Table 1 in the Appendix.} when applying these methods introduced above. Overall, there is no outstanding method that excels in all four evaluation dimensions across the five open-source models tested. For instance, despite the \textit{base} prompting method achieving the highest output quality on all models (red color is close to 5 in all models), 
it falls short in terms of diversity metrics when compared to other methods. The shortcoming in diversity is particularly notable in the case of \texttt{Mistral-7B-instruct-v0.3}: with an average output quality rating of 4.32 in \textit{base} prompting, it merely achieves an output diversity score of 0.33 and an average target domain cluster number of 1.97 on three datasets (Table 1 in the Appendix describes this numbers). In comparison, when applying the Top-$k$ method  (green color in Figure \ref{fig:radar_chart}) on the \texttt{Mistral-7B-instruct-v0.3}, with an average analogy quality of 4.21, it achieves an average diversity score of 0.53 as well as the average number of target domain clusters of 2.27, indicating a compromise of diversity in the \textit{base} prompting method when compared with probabilistic steering methods such as Top-$k$. 

In addition to the lack of a dominating method that excels in all evaluation metrics, we also observe that different models exhibit unstable rankings of these approaches even when evaluated on a single metric. Taking the diversity score metric (indicated at the top of the circle in Figure \ref{fig:radar_chart}) as an example, the best performing methods in enhancing diversity for \texttt{Llama-3.1-8B-instruct} are Top-$p$ and G2 (which belong to the probabilistic and Logistic steering category), both achieving an average score of 0.61 on all three datasets. However, when the same two methods are applied to \texttt{Qwen-3-8B}, they only achieves a diversity score of 0.28 and 0.29, respectively. In comparison, the entropy-gated steering method achieves an average performance of 0.39 in diversity score. 
\section{Detailed Analysis}
\label{sec:analysis}

\begin{figure*}[h]
    \centering
    \begin{minipage}[b]{0.27\textwidth}
        \centering
        \includegraphics[width=\linewidth]{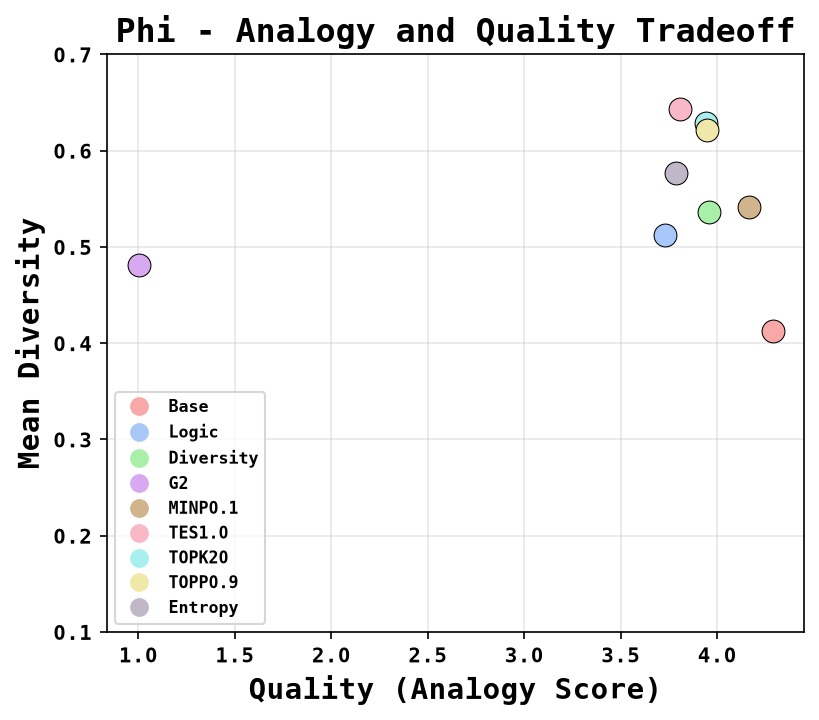}
    \end{minipage}
    \hfill
    \begin{minipage}[b]{0.72\textwidth}
        \centering
        \includegraphics[width=\linewidth]{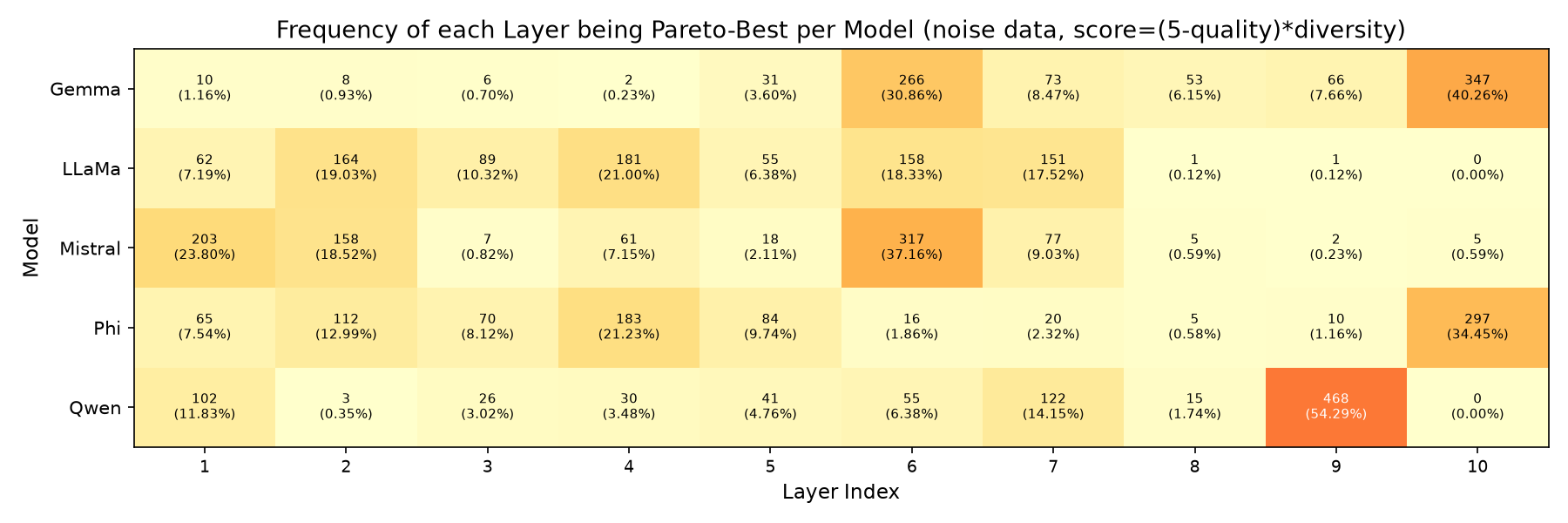}
    \end{minipage}
    \caption{(a) Scatter plot of the Pareto frontier across all diversity enhancing frameworks using \texttt{Phi-4-mini-instruct}. (b) Heatmap of each layer group being selected as Pareto-best under layer-wise noise perturbation across different model families.}
    \label{fig:analysis}
\end{figure*}

In this section, we provide an in-depth analysis, aiming to account for the observations of divergent performance across these diversity-enhancing methods elaborated in the previous section. Specifically, we show that (1) there exists a tradeoff between the output diversity and output quality, and (2) the misalignment between intervention position and the models' analogy-making region may account for the unstable performance in analogy making.


\paragraph{Diversity-quality tradeoff in analogy making.} \hspace{-2mm} Following the observations in the radar chat (Figure \ref{fig:radar_chart} ) showing that the increase in analogy diversity is often accompanied by a decrease in output quality, we further investigate this tradeoff by plotting the Pareto frontier of each model. Specifically, we collect the average performance across three datasets on diversity and quality metrics and make a scatter plot of these methods. Due to space limitations, we only showcase the scatter plot for \texttt{Phi-4-mini-instruct} in Figure \ref{fig:analysis} (a). When compared with the base prompting (light red dot in Figure \ref{fig:analysis} (a)) that yields analogies with an average quality of 4.28, the most diverse method, Top-$\eta \sigma$, (light pink dot in Figure \ref{fig:analysis} (a)) only yields an average quality of 3.80, indicating a compromise in the diversity enhancing approaches in output quality.
Additionally, we also observe a clustering effect of those diversity-enhancing approaches on the Pareto frontier. This further indicates an absence of a dominating approach in improving models' analogy diversity meanwhile maintaining high generation quality. 


\paragraph{Misalignment between intervention position and the analogy-making region in LLMs.} \hspace{-2mm} Given the observed tradeoff between diversity and quality, one might also be curious about whether the internal computations governing analogy generation are confined to particular regions of the model, regardless of the model family. To probe this question, we take inspiration from exiting works in causal perturbation analysis \cite{todd2024function} and introduce the following method that aims to identify the most fragile regions when perturbed with looped computation, a novel perturbation method suggested in \cite{fan2026bridginggaplatentexplicit}.
Specifically, we blur the hidden states via looped inference and compute the indirect effect, a mainstream method introduced by \cite{todd2024function} to identify the most important layers within an LLM. In our case, we define the metric of indirect effect to be the combined score of $(5-\text{quality}) \times \text{diversity}$. The most fragile regions are thus identified by maximizing the metric when replaced with perturbations, meaning that these layers are least informative yet diverse in analogy making. Then, we aggregate, for each model, the frequency with which each layer is selected across all evaluation samples. Figure~\ref{fig:analysis} (b) visualizes these frequencies as a heatmap, revealing the layer positions that are most frequently identified as Pareto-optimal under our perturbation analysis.

From the resulting heatmap, we observe a fluid and divergent span of layer regions that are sensitive to analogy making across different models. For instance, \texttt{llama} and \texttt{mistral} models have a more widespread span of layers (the 20th percentile to the 70th percentile among layer depth) when perturbed with noise, while for \texttt{phi-4-mini} and \texttt{qwen-3}, the sensitive regions are concentrated between certain layer positions (the 90th percentile for \texttt{qwen-3} and the 100th percentile for \texttt{phi-4-mini}). The wide span along with the fluid regions across different models that are sensitive to analogy making also echos with the insights from \cite{minegishi2026arintransformers}, where they argue that analogical reasoning is an ability emerging from the late phase of model internals. In contrast to this observation, however, all steering methods that influence output diversity operate on the logits that are obtained from the last layer. The misalignment between the sensitive layer regions and these perturbation methods thus offer a potential explanation for the unstable performance of methods when applied to different LLMs.


\section{Conclusion and Outlook}
\label{sec:conclusion}

We present a systematic study of output diversity in LLM-based analogy making across ten state-of-the-art open- and closed-source LLMs. Our results indicate  the phenomenon of domain homogeneity from both intra-output and inter-model perspectives, where models repeatedly generate analogies grounded in a narrow set of target domains, a concerning issue that bears the risk of undermining AI4S creativity. Further evaluation on various inference-time diversity-enhancing approaches from three categories shows that their effectiveness is generally limited and strongly model-dependent, and there exists a trade-off in which gains in analogy making diversity are often accompanied by degradation in analogy making quality. Our further analysis also provides a potential explanation via layer perturbation, showing a misalignment between the intervention location of these methods and the model’s sensitive region in analogy making. 

To the best of our knowledge, this work is among the first to systematically investigate output diversity in open-ended LLM analogy generation. Our findings may help inform future LLM model designs that are more diversity-aware in analogy making.

\clearpage
\bibliography{reference}
\bibliographystyle{icml2026}

\clearpage
\appendix

\newenvironment{promptbox}[2][]{%
    \begin{tcolorbox}[
        enhanced,
        breakable,
        colback=gray!5,
        colframe=gray!60,
        coltitle=black,
        fonttitle=\bfseries\small,
        title={#2},
        fontupper=\ttfamily\small,
        left=4pt, right=4pt, top=3pt, bottom=3pt,
        boxrule=0.4pt,
        arc=1mm,
        width=\columnwidth, 
        #1
    ]
}{%
    \end{tcolorbox}
}

\section{Appendix}
\label{sec:appendix}

\appendix

\section{Experimental Setup}
\label{appendix:experimental_setup}

\paragraph{Datasets}
Experiments are conducted on three analogy datasets: \textbf{AB} (\texttt{AnaloBench\_S1}), containing 340 samples; \textbf{MA} (\texttt{metaphoric\_analogies}), containing 262 samples; and \textbf{MUNCH} (\texttt{munch\_metaphors}), containing 300 samples. For each source sentence, every model generates $K=10$ candidate analogies.

\paragraph{Evaluated Models}
We evaluate ten large language models (LLMs), comprising five proprietary and five open-weight models. The proprietary models include Claude Sonnet (\texttt{anthropic/claude-sonnet-4.5}), GPT-5.2 (\texttt{openai/gpt-5.2}), Gemini 2.5 Pro (\texttt{google/gemini-2.5-pro}), Gemini 3.1 Pro (\texttt{google/gemini-3.1-pro-preview}), and Grok 4.5 (\texttt{x-ai/grok-4.5}). The open-weight models consist of Phi-4 (\texttt{microsoft/Phi-4-mini-instruct}), Mistral (\texttt{mistralai/Mistral-7B-Instruct-v0.3}), Gemma 2 (\texttt{google/gemma-2-9b-it}), Qwen 3 (\texttt{Qwen/Qwen3-8B}), and Llama 3.1 (\texttt{meta-llama/Llama-3.1-8B-Instruct}). We access the proprietary models via the OpenRouter API, while the open-weight models are executed locally in FP16 precision using the Hugging Face \texttt{transformers} library
\section{Experiment 1 (a): Analogy Generation}
\label{appendix:exp1a}

The objective of this experiment is to evaluate both the diversity and analogy quality of model-generated sentences under different prompting and decoding strategies.

Unless otherwise specified, all models use standard sampling with a temperature of $T=0.7$ and generate $K=10$ candidate analogies for each input sentence. We evaluate three prompting strategies corresponding to the Base, Diversity, and Logic prompts described in the Prompts section.

To further investigate the diversity--quality trade-off, we additionally evaluate a higher temperature ($T=1.5$) using Base Prompt combined with four decoding strategies: Top-$k$, Top-$p$, Top-$\eta\sigma$, and Min-$p$.

\subsection{Evaluation Pipeline}

\paragraph{Generation}
For every model and decoding configuration, a JSON file is produced containing the source sentences together with the $K$ generated analogies.

\paragraph{Diversity Metrics}

For each input sample, a set of analogies is generated and evaluated using three complementary diversity metrics, which capture semantic, lexical, and vocabulary-level variation. The final diversity score is computed as a weighted combination of these metrics:
\[
D_{\mathrm{total}}
=
0.5D_{\mathrm{cos}}
+
0.25D_{\mathrm{BLEU}}
+
0.25D_{\mathrm{EAD}}.
\]

\textbf{Average Cosine Distance ($D_{\mathrm{cos}}$).}
Each generated analogy is embedded using the SentenceTransformer model \texttt{all-MiniLM-L6-v2}. Pairwise cosine distances are computed between all valid analogy embeddings for each sample, and their average is used as the semantic diversity score. Higher values indicate greater semantic dispersion among the generated analogies. Samples containing fewer than two analogies receive a score of zero.

\textbf{Div-BLEU ($D_{\mathrm{BLEU}}$).}
Lexical diversity is measured using Self-BLEU. Each analogy is treated once as the candidate while all remaining analogies serve as references. Sentence-level BLEU is computed using SacreBLEU with exponential smoothing, and Self-BLEU is obtained by averaging across all candidates. The diversity score is then defined as
\[
D_{\mathrm{BLEU}}
=
1-\mathrm{Self\mbox{-}BLEU},
\]
such that higher values correspond to lower lexical overlap.

\textbf{Expectation-Adjusted Distinctness ($D_{\mathrm{EAD}}$).}
To measure vocabulary diversity while correcting the length bias of the original Distinct-$n$ metric, we adopt Expectation-Adjusted Distinctness (EAD).

For each generation method, the reported diversity statistics correspond to the mean, median, and standard deviation of $D_{\mathrm{total}}$ across all evaluated samples.

\paragraph{Analogy Quality}

The quality of every generated analogy is independently assessed by a judge LLM using the Analogy Quality Scoring Prompt described in the prompt section in Appendix. The judge receives the original sentence, the generated analogy, and a formal definition of analogy based on domain mapping theory, and assigns an integer score from 1 (poor) to 5 (excellent) according to the validity of the mapping, semantic coherence, and overall meaningfulness. To ensure deterministic evaluation, inference is performed with temperature zero, and only the integer score is extracted from the model output. Responses outside the valid range are discarded.

For each input sample, the quality score is computed as the mean of the scores assigned to its generated analogies. The quality of a generation method (i.e., a specific combination of model, prompt, and decoding strategy) is then reported as the mean and standard deviation of these sample-level scores across the entire evaluation set.

\begin{figure*}[ht]
    \centering
    \includegraphics[width=\textwidth]{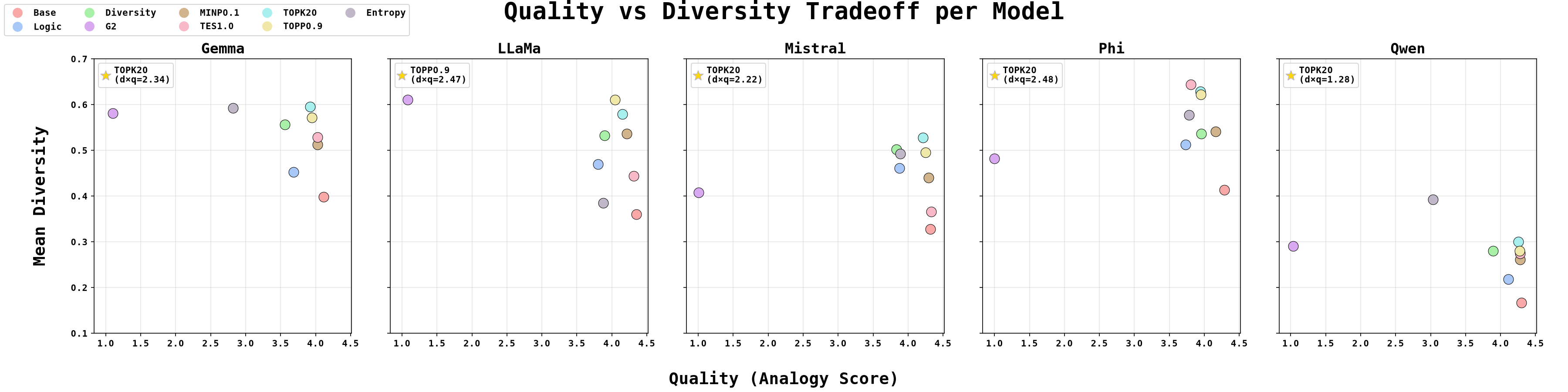}
    \caption{Quality--diversity tradeoff for each evaluated language model. Each point corresponds to a prompting or decoding strategy, positioned according to its mean analogy quality and mean diversity. The highlighted star denotes the selected operating point, determined by maximizing the combined objective $\mathrm{Quality} \times \mathrm{Diversity}$.}
    \label{fig:quality_vs_diversity_scatter}
\end{figure*}

\section{Experiment 1 (b): Domain Extraction}
\label{appendix:exp1b}

This experiment measures the diversity of conceptual domains employed by language models during analogy generation. For each generated analogy, GPT-4o-mini ($T=0.0$) extracts its source and target domains in JSON format using the Domain Extraction Prompt described in the Prompts section.

\paragraph{Domain Diversity}

\begin{figure}[t]
    \centering
    \includegraphics[width=\linewidth]{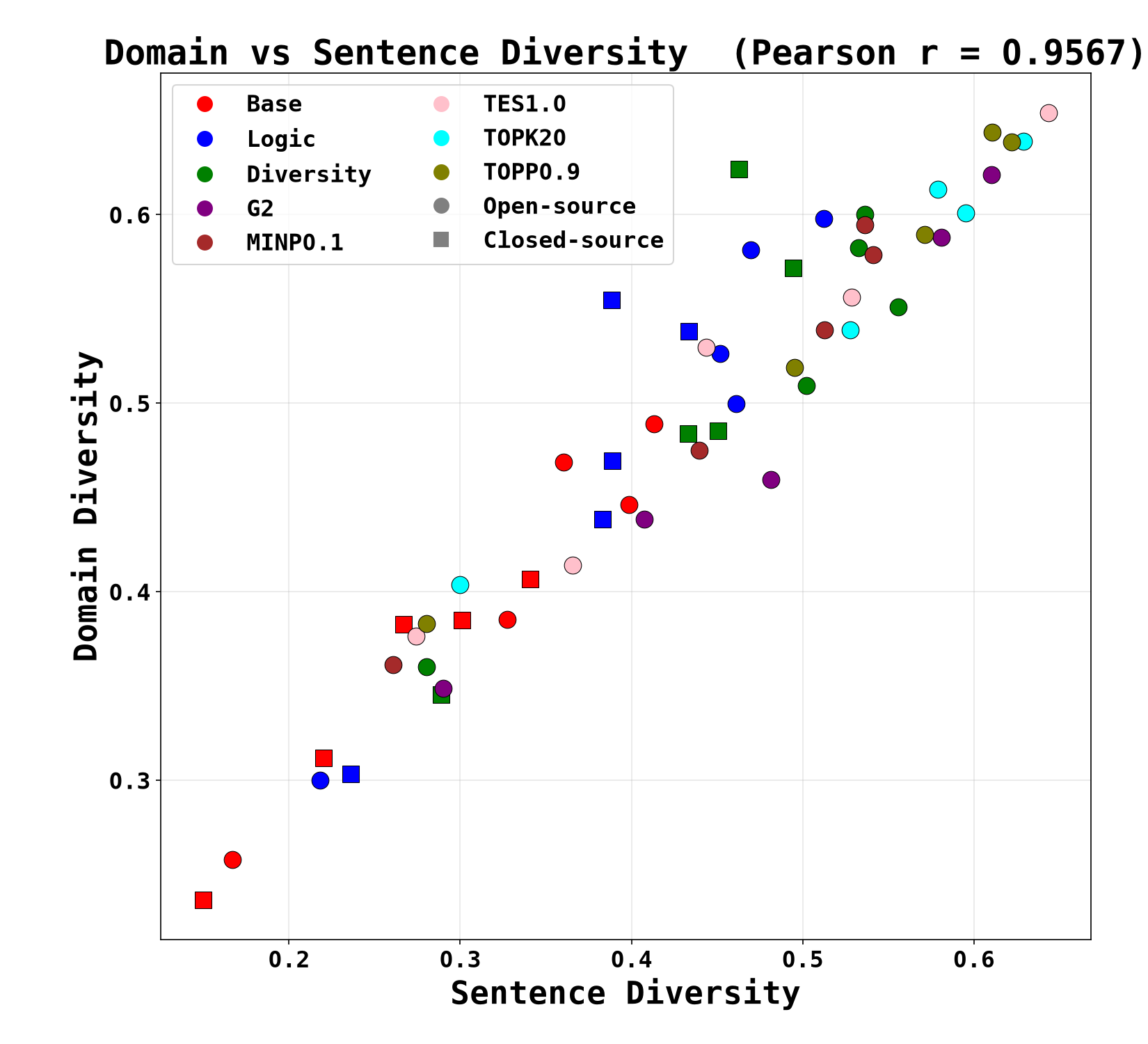}
    \caption{Relationship between sentence diversity and domain diversity across all evaluated generation methods. Each point corresponds to a model--method configuration, with colors indicating the prompting strategy and marker shapes distinguishing open-source and closed-source models.}
    \label{fig:domain_vs_sentence_diversity}
\end{figure}

The extracted target-domain labels are treated as text samples and evaluated using the same diversity metrics described in Experiment~1(a). This allows us to quantify the semantic diversity of the conceptual domains independently of the linguistic variation in the generated analogies. Figure~\ref{fig:domain_vs_sentence_diversity} compares domain diversity with sentence diversity across all evaluated generation methods. The strong positive correlation indicates that methods producing more diverse analogies also tend to explore a broader range of conceptual domains.

\section{Experiment 2: MAUVE Similarity}
\label{appendix:exp2}

This experiment measures distributional similarity between proprietary
language models using MAUVE.

Only generations produced using the Base prompt in the Prompts section with $T=0.7$ are considered. Outputs from every model are pooled across all datasets before computing
pairwise MAUVE scores. GPT-2 Large hidden representations are used for featurization, and the resulting pairwise similarities are presented as a heatmap.

\section{Experiment 3: Memorization Test}
\label{appendix:exp3}
This experiment investigates whether proprietary models have memorized examples from the AnaloBench dataset.

For every prefix ratio

\[
r \in \{0.1,0.2,0.4,0.8\},
\]

the first $r$ fraction of the reference completion is provided to the model. 

Using the Base Prompt described in Prompts section, models are instructed to reproduce the continuation from the provided prefix using few-shot examples. Performance is evaluated using Token Accuracy, Character Accuracy, Exact Match (EM), and Contains Accuracy. Higher Exact Match scores indicate stronger evidence of memorization.

\section{Experiment 4: LLM-based Clustering}
\label{appendix:exp4}

This experiment investigates the semantic organization of generated
analogies through LLM-based clustering.

For every source sentence, the corresponding $K$ generated analogies are submitted to GPT-4o-mini, which groups them into semantic categories represented as JSON objects using the Clustering prompt described in the Prompts section in Prompts section. The prompt instructs the model to cluster analogies according to their conceptual domains while avoiding lexical variations and synonym duplication.

\paragraph{Cluster-Linkage Analysis}
Cluster embeddings are computed using SentenceTransformer (\texttt{all-MiniLM-L6-v2}). For each pair of clusters within a sample, we compute the single, complete, and average linkage. The linkage score for a given method is obtained by averaging the corresponding linkage values across all samples. Samples containing only a single cluster are assigned a linkage score of 0. Additionally, we report the average number of semantic clusters per source sentence.

\section{Experiment 5: Fragile Region Identification}
\label{appendix:exp5}

\paragraph{Noise Blur}
This experiment investigates whether analogy generation is associated with specific regions of transformer models or is distributed across the network. For each input sample, the model generates $K$ analogies by applying hidden-state perturbations independently at transformer layers located at 10\% depth intervals, from 10\% to 100\% of the network. The generated analogies are evaluated using the same diversity and quality metrics described in Experiment~1(a), yielding a diversity score and an analogy quality score for every layer and sample. These scores are subsequently used to identify the layers most strongly associated with the analogy-making process through a perturbation-based analysis.

\paragraph{Layer Evaluation}
For each input sample, every transformer layer is represented as a point in the quality--diversity space, where quality is minimized and diversity is maximized. The Pareto frontier is then computed by retaining only the non-dominated layers. Among the Pareto-optimal candidates, a single layer is selected by maximizing the combined objective
\[
(5-\mathrm{Quality}) \times \mathrm{Diversity},
\]
which favors layers achieving both low-quality and diverse analogy generation. This procedure is repeated independently for every sample. Finally, for each language model, we report the frequency with which each layer is selected as the optimal Pareto layer across the evaluation set, providing an estimate of the model regions that are most consistently sensitive to hidden-state perturbations during analogy generation.


\onecolumn
\subsection{Complete Results}
This section presents the complete experimental results in Table~\ref{tab:main_results}.

\begin{table*}[ht]
    \centering
    \small 
    \renewcommand{\arraystretch}{1.2}
    \setlength{\tabcolsep}{4pt} 
    \scalebox{0.7}{
    \begin{tabularx}{25cm}{
        >{\centering\arraybackslash}p{2.5cm} 
        >{\centering\arraybackslash}p{2.5cm} 
        >{\centering\arraybackslash}p{1.5cm} 
        >{\centering\arraybackslash}p{1.3cm}
        >{\centering\arraybackslash}p{1.3cm}
        >{\centering\arraybackslash}p{1.3cm}
        >{\centering\arraybackslash}p{1.5cm} 
        >{\centering\arraybackslash}p{1.3cm}
        >{\centering\arraybackslash}p{1.3cm}
        >{\centering\arraybackslash}p{1.3cm}
        >{\centering\arraybackslash}p{1.5cm} 
        >{\centering\arraybackslash}p{1.3cm}
        >{\centering\arraybackslash}p{1.3cm}
        >{\centering\arraybackslash}p{1.3cm}
    }
    \toprule
    \multirow{2}{*}{\textbf{Model Name}} & \multirow{2}{*}{\textbf{Method}} & \multicolumn{4}{c}{\textbf{AB}} & \multicolumn{4}{c}{\textbf{MA}} & \multicolumn{4}{c}{\textbf{MUNCH}} \\
    \cmidrule(lr){3-6} \cmidrule(lr){7-10} \cmidrule(lr){11-14}
    &  & Div. $\uparrow$ & Qual. $\uparrow$ & Clus. Linkage & \#Clus. & Div. $\uparrow$ & Qual. $\uparrow$ & Clus. Linkage & \#Clus. & Div. $\uparrow$ & Qual. $\uparrow$ & Clus. Linkage & \#Clus. \\
    \midrule

    \multirow{7}{*}{\parbox{2.2cm}{\texttt{Llama-3.1-}\\ \texttt{8B-instruct}}} 
    & Std. Prompt & 0.355 & \textbf{\underline{4.66}} & 0.351 & 2.10 & 0.378 & \underline{4.06} & 0.421 & 2.40 & 0.348 & \underline{4.33} & 0.286 & 1.97 \\
    \cdashline{2-14}
    & Div. Prompt  & 0.564 & 4.18 & \underline{0.639} & \textbf{\underline{3.24}} & 0.495 & 3.71 & \underline{0.577} & \underline{2.96} & 0.538 & 3.80 & \underline{0.590} & \underline{3.28} \\
    & Top-$k$ & 0.584 & 4.46 & 0.432 & 2.18 & 0.582 & 3.83 & 0.550 & 2.68 & 0.571 & 4.18 & 0.395 & 2.16 \\
    & Top-$p$  & \underline{0.613} & 4.40 & 0.450 & 2.24 & 0.614 & 3.70 & 0.571 & 2.70 & 0.604 & 4.04 & 0.446 & 2.23 \\ 
    & Top-$\eta\sigma$  & 0.446 & 4.64 & 0.359 & 2.09 & 0.455 & 4.01 & 0.459 & 2.57 & 0.430 & 4.30 & 0.347 & 2.18 \\
    & Min-$p$  & 0.542 & 4.53 & 0.435 & 2.31 & 0.543 & 3.89 & 0.502 & 2.63 & 0.523 & 4.22 & 0.394 & 2.18 \\
    & Entropy  & 0.393 & 4.36 & 0.376 & 2.17 & 0.465 & 3.54 & 0.526 & 2.70 & 0.384 & 3.88 & 0.395 & 2.33 \\
    & G2  & 0.601 & 1.07 & 0.474 & 2.03 & \underline{0.620} & 1.10 & 0.503 & 2.06 & \underline{0.610} & 1.07 & 0.385 & 1.73 \\

    \noalign{\vskip 1pt}\midrule\noalign{\vskip 1pt}

    \multirow{7}{*}{\texttt{Gemma2-9B-it}} 
    & Std. Prompt & 0.374 & \underline{4.44} & 0.420 & 2.15 & 0.422 & \underline{3.77} &0.493 & 2.38 & 0.399 & \underline{4.13} & 0.410 & 2.27 \\
    \cdashline{2-14}
    & Div. Prompt  & 0.564 & 3.85 & 0.690 & \underline{3.09} & 0.514 & 3.46 & 0.582 & 2.93 & 0.590 & 3.37 & 0.686 & \textbf{\underline{3.54}} \\
    & Top-$k$ & 0.584 & 4.26 & 0.536 & 2.53 & 0.609 & 3.57 & 0.606 & 2.91 & 0.593 & 3.93 & 0.509 & 2.62 \\
    & Top-$p$  & 0.549 & 4.29 & 0.511 & 2.51 & 0.592 & 3.59 & 0.581 & 2.87 & 0.572 & 3.95 & 0.503 & 2.72 \\ 
    & Top-$\eta\sigma$  & 0.506 & 4.36 & 0.496 & 2.32 & 0.556 & 3.66 & 0.561 & 2.71 & 0.523 & 4.06 & 0.493 & 2.66 \\
    & Min-$p$  & 0.493 & 4.38 & 0.471 & 2.38 & 0.534 & 3.67 & 0.560 & 2.61 & 0.511 & 4.03 & 0.454 & 2.60 \\
    & Entropy  & \underline{0.589} & 3.40 & \textbf{\underline{0.721}} & 2.82 & \underline{0.619} & 2.88 & \textbf{\underline{0.751}} & \underline{2.97} & 0.592 & 2.82 & \textbf{\underline{0.743}} & 2.77 \\
    & G2  & 0.570 & 1.08 & 0.461 & 2.05 & 0.573 & 1.12 & 0.506 & 2.03 & \underline{0.599} & 1.09 & 0.466 & 1.90 \\
    
    \noalign{\vskip 1pt}\midrule\noalign{\vskip 1pt}

    & Std. Prompt & 0.417 & \underline{4.57} & 0.399 & 2.30 & 0.423 & \underline{3.97} & 0.450 & 2.33 & 0.399 & \underline{4.31} & 0.325 & 2.15 \\
    \cdashline{2-14}
    \multirow{7}{*}{\parbox{2.2cm}{\texttt{Phi-4-mini-}\\ \texttt{instruct}}} 
    & Div. Prompt  & 0.543 & 4.31 & \underline{0.596} & \underline{3.09} & 0.515 & 3.71 & \underline{0.607} & \textbf{\underline{3.28}} & 0.550 & 3.85 & \underline{0.610} & \underline{3.48} \\
    & Top-$k$ & 0.641 & 4.19 & 0.520 & 2.48 & 0.628 & 3.64 & 0.556 & 2.70 & 0.616 & 4.00 & 0.453 & 2.34 \\
    & Top-$p$  & 0.629 & 4.24 & 0.498 & 2.43 & 0.628 & 3.60 & 0.570 & 2.75 & 0.609 & 4.00 & 0.448 & 2.42 \\ 
    & Top-$\eta\sigma$  & \textbf{\underline{0.659}} & 4.03 & 0.523 & 2.57 & \textbf{\underline{0.638}} & 3.49 & 0.574 & 2.86 & \textbf{\underline{0.632}} & 3.89 & 0.494 & 2.49 \\
    & Min-$p$  & 0.545 & 4.45 & 0.426 & 2.34 & 0.552 & 3.83 & 0.517 & 2.74 & 0.526 & 4.20 & 0.379 & 2.29 \\
    & Entropy  & 0.603 & 4.24 & 0.493 & 2.01 & 0.560 & 3.62 & 0.477 & 2.16 & 0.577 & 3.79 & 0.477 & 2.04 \\
    & G2  & 0.487 & 1.00 & 0.128 & 1.27 & 0.481 & 1.00 & 0.167 & 1.42 & 0.476 & 1.00 & 0.105 & 1.21 \\
    
    \noalign{\vskip 1pt}\midrule\noalign{\vskip 1pt}
    & Std. Prompt & 0.325 & 4.56 & 0.317 & 1.94 & 0.351 & 4.02 & 0.394 & 2.18 & 0.307 & 4.37 & 0.230 & 1.78 \\
    \cdashline{2-14}
    \multirow{7}{*}{\parbox{2.2cm}{\texttt{Mistral-7B-}\\ \texttt{instruct-v0.3}}} 
    & Div. Prompt  & 0.518 & 4.16 & \underline{0.604} & \underline{3.06} & 0.468 & 3.68 & 0.542 & \underline{2.95} & \underline{0.520} & 3.66 & \underline{0.606} & \textbf{\underline{3.54}} \\
    & Top-$k$ & \underline{0.538} & 4.49 & 0.404 & 2.24 & 0.545 & 3.90 & 0.505 & 2.57 & 0.500 & 4.23 & 0.302 & 2.00 \\
    & Top-$p$  & 0.497 & 4.53 & 0.382 & 2.13 & 0.518 & 3.93 & 0.462 & 2.41 & 0.470 & 4.29 & 0.336 & 2.05 \\ 
    & Top-$\eta\sigma$  & 0.368 & \underline{4.58} & 0.336 & 1.99 & 0.386 & \textbf{\underline{4.04}} & 0.398 & 2.22 & 0.343 & \textbf{\underline{4.38}} & 0.251 & 1.83 \\
    & Min-$p$  & 0.441 & 4.55 & 0.347 & 2.06 & 0.460 & 4.01 & 0.454 & 2.40 & 0.417 & 4.32 & 0.306 & 1.91 \\
    & Entropy  & 0.515 & 4.20 & 0.387 & 1.88 & \underline{0.565} & 3.65 & \underline{0.549} & 2.49 & 0.492 & 3.89 & 0.359 & 1.97 \\
    & G2  & 0.389 & 1.00 & 0.148 & 1.38 & 0.416 & 1.01 & 0.215 & 1.58 & 0.418 & 1.00 & 0.131 & 1.30 \\

    \noalign{\vskip 1pt}\midrule\noalign{\vskip 1pt}
    & Std. Prompt & 0.160 & \underline{4.61} & 0.203 & 1.59 & 0.172 & \underline{4.03} & 0.229 & 1.65 & 0.170 & \underline{4.26} & 0.206 & 1.63 \\
    \cdashline{2-14}
    \multirow{7}{*}{\parbox{2.2cm}{\texttt{Qwen3-8B}}} 
    & Div. Prompt  & 0.282 & 4.21 & \underline{0.407} & \underline{2.24} & 0.263 & 3.72 & 0.363 & 2.03 & 0.296 & 3.75 & \underline{0.398} & \underline{2.31} \\
    & Top-$k$ & 0.285 & 4.56 & 0.311 & 2.06 & 0.314 & 3.98 & 0.361 & 2.19 & 0.300 & 4.22 & 0.309 & 2.14 \\
    & Top-$p$  & 0.266 & 4.60 & 0.305 & 2.01 & 0.294 & 4.01 & \underline{0.372} & \underline{2.23} & 0.281 & 4.21 & 0.304 & 2.07 \\ 
    & Top-$\eta\sigma$  & 0.263 & 4.58 & 0.290 & 1.89 & 0.285 & 4.02 & 0.353 & 2.11 & 0.274 & 4.24 & 0.305 & 2.05 \\
    & Min-$p$  & 0.252 & 4.60 & 0.286 & 1.93 & 0.269 & 4.00 & 0.317 & 2.00 & 0.261 & 4.24 & 0.306 & 2.08 \\
    & Entropy  & \underline{0.377} & 3.47 & 0.334 & 1.81 & \underline{0.398} & 2.95 & 0.362 & 1.78 & \underline{0.394} & 3.04 & 0.300 & 1.63 \\
    & G2  & 0.287 & 1.03 & 0.177 & 1.42 & 0.292 & 1.04 & 0.222 & 1.61 & 0.290 & 1.03 & 0.174 & 1.37 \\

    \bottomrule
    \end{tabularx}}
    \caption{Evaluation results on three datasets using different generation perturbation methods. AB, MA stands for AnaloBench \cite{analobench} and Meterphoric Analogy \cite{tong-etal-2024-metaphor}, respectively. The best results among the perturbation methods are \underline{underlined}, and the best results across different models are \textbf{bolded}. }
    \label{tab:main_results}
\end{table*}

\section{Prompts}
\label{appendix:prompts}

This section presents all prompt templates used throughout the experiments. In every template, text enclosed in \texttt{\{curly\_braces\}} denotes a placeholder that is replaced with the corresponding input at runtime.

\begin{promptbox}{Generation Base Prompt}
\label{prompt:base}
Generate a sentence that is analogous to the sentence below.
Directly output the answer with no explanation.

Sentence: \{source\_story\}\\
Answer:
\end{promptbox}

\begin{promptbox}{Generation Logic Prompt}
\label{prompt:logic}
Generate a sentence that follows the same relational structure
of the sentence below, but in a completely different domain.
Directly output the sentence with no explanation.

Sentence: \{source\_story\}\\
Answer:
\end{promptbox}

\begin{promptbox}{Generation Diversity Prompt}
\label{prompt:diversity}
Generate a sentence that is analogous to the sentence below.
Be as creative and diverse as possible in your choice of domain.
Directly output the answer with no explanation.

Sentence: \{source\_story\}\\
Answer:
\end{promptbox}

\begin{promptbox}{Analogy Quality Scoring Prompt}
\label{prompt:quality}
Read the sentence and the generated analogy, and give it a score between 1 and 5 based on whether the generated analogy is a valid analogy for the sentence.\\[0.5em]

The official definition of analogy can be found below:\\[0.5em]
"An analogy is composed of object mappings and shared relations. Object mappings establish correspondences between equivalent objects across different domains, while shared relations represent equivalent relationships between these objects in both domains."\\[0.5em]

Think carefully before making your verdict. When assigning the score, also evaluate whether the generated analogy is coherent and meaningful.\\[0.5em]

Sentence: \{sentence\}\\
Generated Analogy: \{analogy\}\\[0.5em]

Respond ONLY with a single integer from 1 to 5.
\end{promptbox}

\begin{promptbox}{Domain Extraction Prompt}
\label{prompt:domain-extraction}
You are analyzing an analogy to identify its conceptual structure it establishes.\\
An analogy maps objects from a source domain to corresponding objects in a different target domain while preserving their relationships.\\[0.5em]

**Input sentence:**\\
\{\{input\_sentence\}\}\\[0.5em]

**Provided analogy:**\\
\{\{input\_analogy\}\}\\[0.5em]

**Your task:**\\
Analyze the provided analogy and identify the conceptual domains involved.
Do NOT generate a new analogy or modify the provided one. Only extract the conceptual structure already present.\\[0.5em]

**Rules for extracting domains:**\\
1. Granularity: Use broad, standard fields of activity, industries, or areas of knowledge.\\
2. No Specific Entities: Do NOT use specific events, projects, or historical names.\\
3. Format: Use lowercase, singular noun phrases. Max 3 words per domain.\\
4. Language: Output the domains in English.\\[0.5em]

**Return a JSON object with:**\\[0.5em]

1. "source\_domain": The primary conceptual or real-world domain represented in the original input sentence.\\
2. "target\_domain": The conceptual or real-world domain represented in the provided analogy.\\[0.5em]

Return ONLY the JSON object, with no additional text.
\end{promptbox}

\begin{promptbox}{Clustering Prompt}
\label{prompt:clustering}
You are given \{K\} sentences. Your task is to group them into semantic categories based on their underlying conceptual domain.\\[0.5em]

**Task**\\
Analyze the meaning of each sentence and assign each sentence to exactly one category representing its primary conceptual domain.\\[0.5em]

**Category Guidelines**\\
* Categories must represent broad semantic concepts, not the exact wording used in the sentences.\\
* Normalize synonyms and closely related concepts into the same category.\\
* Do not create separate categories for lexical variations that refer to the same underlying domain.\\
  Examples:\\
    - "recipe creation", "recipe invention", "recipe development" -> same category\\
    - "car repair" and "vehicle maintenance" -> same category\\
    - "writing a book" and "book creation" -> same category\\
* Prefer abstract domain names that describe the shared concept.\\
* Category names should not simply copy phrases from individual sentences.\\[0.5em]

**Classification Rules**\\
* Analyze the meaning of each sentence, not just keywords.\\
* Identify the main subject/domain of the sentence.\\
* Group sentences together when they concern the same conceptual area, even if they use different wording or describe different scenarios.\\
* If a sentence is an analogy, classify it according to the domain of the objects being compared, not according to the analogy structure.\\
* Do not group sentences from different domains just because they share similar relationships, emotions, or logical patterns.\\[0.5em]

**Category Creation Rules**\\
* The number of categories is not predefined.\\
* Create a new category only when the underlying conceptual domain is genuinely different.\\
* Before creating a new category, check whether an existing category can describe the sentence after semantic normalization.\\
* Prefer fewer categories when categories would only differ by wording.\\
* Prefer more categories when the real-world domains differ.\\[0.5em]

**Constraints**\\
* Every sentence must belong to exactly one category.\\
* Do not duplicate sentence indices.\\
* Do not create categories that differ only by synonyms.\\[0.5em]

**Input**\\
A numbered list of sentences:\\[0.5em]

\{sentences\}\\[0.5em]

**Output Requirements**\\
Return ONLY a valid JSON object.\\[0.5em]

* Keys must be concise category names representing semantic domains.\\
* Values must be arrays containing the sentence indices belonging to that category.\\
* Do not include explanations or additional text.\\[0.5em]

Example output:\\
\{
  "Recipe Development": [1, 2, 3],
  "Vehicle Maintenance": [4, 5]
\}
\end{promptbox}

\end{document}